\documentclass{article}

\usepackage{hyperref}
\usepackage{url}
\usepackage{booktabs}
\usepackage{amsfonts}
\usepackage{nicefrac}
\usepackage{microtype}
\usepackage{graphicx}
\usepackage{color}
\usepackage{amsmath}

\usepackage[a4paper, total={6.2in, 9.3in}]{geometry}

\usepackage{caption}
\usepackage{subcaption}
\usepackage{verbatim}
\usepackage{multirow}

\usepackage{multirow}
\usepackage{authblk}
\usepackage{rotating}

\usepackage{authblk}

\title{\huge{Visual Interaction Networks}}

\author{Nicholas Watters}
\author{Andrea Tacchetti}
\author{Th\'eophane Weber}
\author{\authorcr Razvan Pascanu}
\author{Peter Battaglia}
\author{Daniel Zoran}
\affil{
    \texttt{\normalsize \{nwatters, atacchet, theophane,} \\
    \texttt{\normalsize razp, peterbattaglia, danielzoran\}@google.com}
}
\affil{DeepMind\\London, United Kingdom}

\date{}

\begin{document}

\maketitle

\begin{abstract}
From just a glance, humans can make rich predictions about the future state of a wide range of physical systems.  On the other hand, modern approaches from engineering, robotics, and graphics are often restricted to narrow domains and require direct measurements of the underlying states. We introduce the \textit{Visual Interaction Network}, a general-purpose model for learning the dynamics of a physical system from raw visual observations. Our model consists of a perceptual front-end based on convolutional neural networks and a dynamics predictor based on interaction networks. Through joint training, the perceptual front-end learns to parse a dynamic visual scene into a set of factored latent object representations. The dynamics predictor learns to roll these states forward in time by computing their interactions and dynamics, producing a predicted physical trajectory of arbitrary length. We found that from just six input video frames the Visual Interaction Network can generate accurate future trajectories of hundreds of time steps on a wide range of physical systems. Our model can also be applied to scenes with invisible objects, inferring their future states from their effects on the visible objects, and can implicitly infer the unknown mass of objects. Our results demonstrate that the perceptual module and the object-based dynamics predictor module can induce factored latent representations that support accurate dynamical predictions. This work opens new opportunities for model-based decision-making and planning from raw sensory observations in complex physical environments.
\end{abstract}

\section{Introduction}\label{S:intro}
Physical reasoning is a core domain of human knowledge \cite{spelke2007core} and among the earliest topics in AI \cite{winograd1971procedures,winston1970learning}. However, we still do not have a system for physical reasoning that can approach the abilities of even a young child. A key obstacle is that we lack a general-purpose mechanism for making physical predictions about the future from sensory observations of the present. Overcoming this challenge will help close the gap between human and machine performance on important classes of behavior that depend on physical reasoning, such as model-based decision-making \cite{battaglia2013simulation}, physical inference \cite{hamrick2016inferring}, and counterfactual reasoning \cite{gerstenberg2012noisy, gerstenberg2014counterfactual}.

We introduce the Visual Interaction Network (VIN), a general-purpose model for predicting future physical states from video data. The VIN is learnable and can be trained from supervised data sequences which consist of input image frames and target object state values. It can learn to approximate a range of different physical systems which involve interacting entities by implicitly internalizing the rules necessary for simulating their dynamics and interactions.

The VIN model is comprised of two main components: a visual encoder based on convolutional neural networks (CNNs) \cite{lecun2015deep}, and a recurrent neural network (RNN) with an interaction network (IN) \cite{battaglia2016interaction} as its core, for making iterated physical predictions. Using this architecture we are able to learn a model which infers object states and can make accurate predictions about these states in future time steps. We show that this model outperforms various baselines and can generate compelling future rollout trajectories.

\subsection{Related work}
One approach to learning physical reasoning is to train models to make state-to-state predictions. The first algorithm using this approach was the ``NeuroAnimator'' \cite{grzeszczuk1998neuroanimator}, which was able to simulate articulated bodies.
Ladicky et al. \cite{ladicky2015data} proposed a learned model for simulating fluid dynamics based on regression forests. Battaglia et al. \cite{battaglia2016interaction} introduced a general-purpose learnable physics engine, termed an Interaction Network (IN), which could learn to predict gravitational systems, rigid body dynamics, and mass-spring systems. Chang et al. \cite{chang2016compositional} introduced a similar model in parallel that could likewise predict rigid body dynamics.

Another class of approaches learn to predict summary physical judgments and produce simple actions from images. There have been several efforts \cite{lerer2016learning,li2016fall} which used CNN-based models to predict whether a stack of blocks would fall. Mottaghi et al. \cite{mottaghi2016newtonian, mottaghi2016happens} predicted coarse, image-space motion trajectories of objects in real images. Several efforts \cite{bhat2002computing, brubaker2009estimating, wu2016physics, wu2015galileo} have fit the parameters of
Newtonian mechanics equations to systems depicted in images and videos, though the dynamic equations themselves were not learned. 
Agrawal et al. \cite{agrawal2016learning} trained a system that learns to move objects by poking.

A third class of methods \cite{bhattacharyya2016long,ehrhardt2017learning,fragkiadaki2015learning,stewart2016label} have been used to predict future state descriptions from pixels. These models have to be tailored to the particular physical domain of interest, are only effective over a few time steps, and use side information such as object locations and physical constraints at test time.

\section{Model}\label{S:model}
\begin{figure}[htb]
  \centering
  \includegraphics[width=0.7\linewidth]{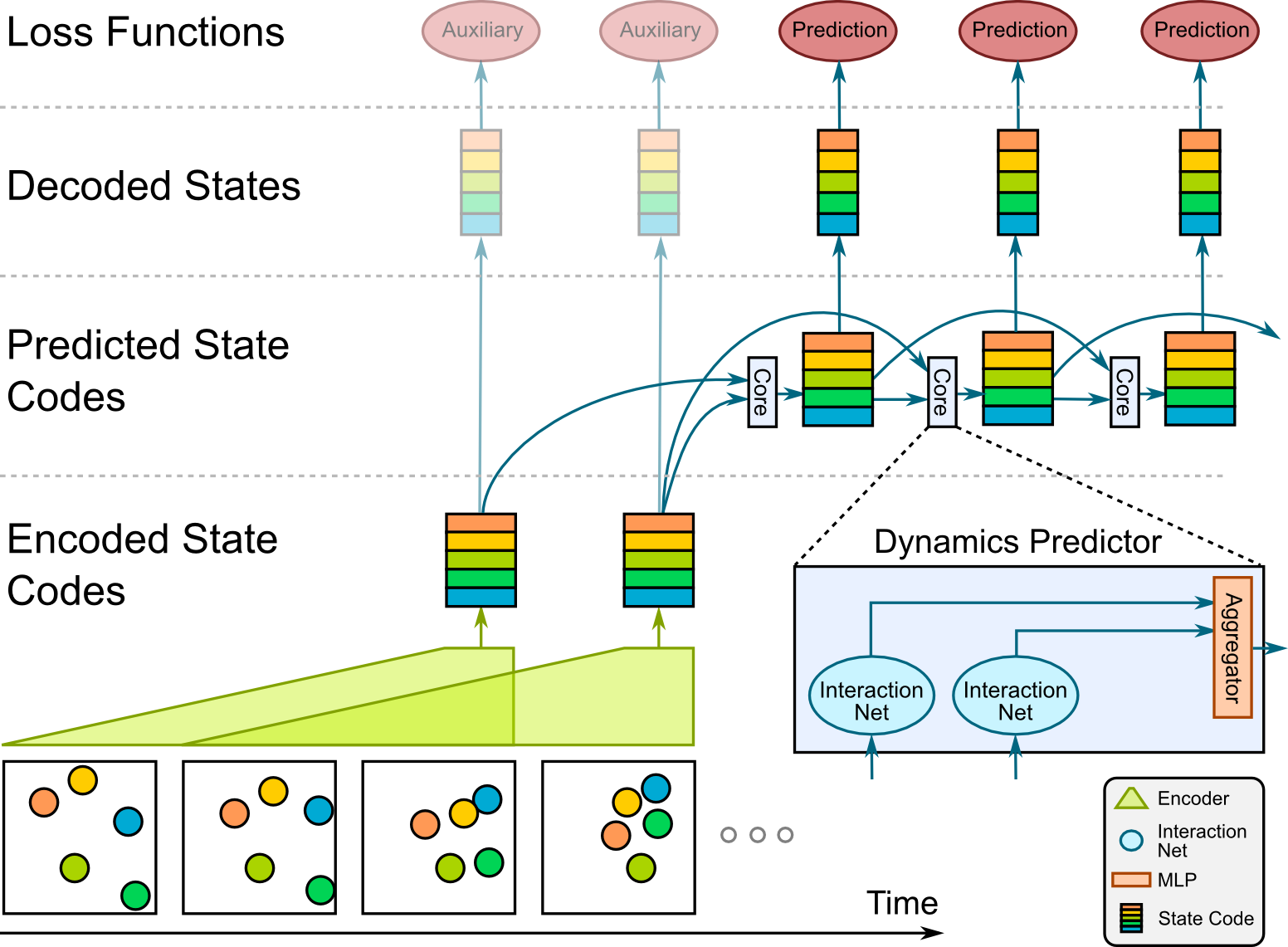}
  \caption{\textbf{Visual Interaction Network:} The general architecture is depicted here (see legend on the bottom right). The visual encoder takes triplets of consecutive frames and produces a state code for the third frame in each triplet. The visual encoder is applied in a sliding window over the input sequence to produce a sequence of state codes. Auxiliary losses applied to the decoded output of the encoder help in training. The state code sequence is then fed into the dynamics predictor which has several Interaction Net cores (2 in this example) working on different temporal offsets. The outputs of these Interaction Nets are then fed into an aggregator to produce the prediction for the next time step. The core is applied in a sliding window manner as depicted in the figure. The predicted state codes are linearly decoded and are used in the prediction loss when training.}
  \label{fig:model_framework}
  \vspace{-1.0em}
\end{figure}

Our Visual Interaction Network (VIN) learns to produce future trajectories of objects in a physical system from only a few video frames of that system.  The VIN is depicted in Figure \ref{fig:model_framework} (best viewed in color), and consists of the following components:
\begin{itemize}
    \item The \textbf{visual encoder} takes a triplet of frames as input and outputs a state code. A state code is a list of vectors, one for each object in the scene. Each of these vectors is a distributed representation of the position and velocity of its corresponding object. We apply the encoder in a sliding window over a sequence of frames, producing a sequence of state codes. See Section \ref{S:encoder} and Figure \ref{fig:frame_pair_encoder} for details.
    
    \item The \textbf{dynamics predictor} takes a sequence of state codes (output from a visual encoder applied in a sliding-window manner to a sequence of frames) and predicts a candidate state code for the next frame. The dynamics predictor is comprised of several interaction-net cores, each taking input at a different temporal offset and producing candidate state codes. These candidates are aggregated by an MLP to produce a predicted state code for the next frame. See Section \ref{S:predictor} and Figure \ref{fig:interaction_net} for details.
    
    \item The \textbf{state decoder} converts a state code to a state. A state is a list of each object's position/velocity vector. The training targets for the system are ground truth states.  See Section \ref{S:decoder} for details.
\end{itemize}

\subsection{Visual Encoder}\label{S:encoder}
The visual encoder is a CNN that produces a state code from a sequence of 3 images. It has a frame pair encoder $E_{\text{pair}}$ shown in Figure \ref{fig:frame_pair_encoder} which takes a pair of consecutive frames and outputs a candidate state code. This frame pair encoder is applied to both consecutive pairs of frames in a sequence of 3 frames. The two resulting candidate state codes are aggregated by a slot-wise MLP into an encoded state code. $E_{\text{pair}}$ itself applies a CNN with two different kernel sizes to a channel-stacked pair of frames, appends constant $x,y$ coordinate channels, and applies a CNN with alternating convolutional and max-pooling layers until unit width and height. The resulting tensor of shape $1\times 1\times (N_{\text{object}} \cdot L_{\text{code}})$ is reshaped into a state code of shape $N_{\text{object}} \times L_{\text{code}}$, where $N_{\text{object}}$ is the number of objects in the scene and $L_{\text{code}}$ is the length of each state code slot. The two state codes are fed into an MLP to produce the final encoder output from the triplet. See the Supplementary Material for further details of the visual encoder model.

One important feature of this visual encoder architecture is its weight sharing given by applying the same $E_{pair}$ on both pairs of frames, which approximates a temporal convolution over the input sequence. Another important feature is the inclusion of constant coordinate channels (an x- and y-coordinate \texttt{meshgrid} over the image), which allows positions to be incorporated throughout much of the processing. Without the coordinate channels, such a convolutional architecture would have to infer position from the boundaries of the image, a more challenging task.\vspace{-0.6em}

\subsection{Dynamics Predictor}\label{S:predictor}
\begin{figure}[htb]
  \centering
  \begin{subfigure}{0.47\textwidth}
  \caption{Frame Pair Encoder}
  \label{fig:frame_pair_encoder}
  \includegraphics[width=1.0\linewidth]{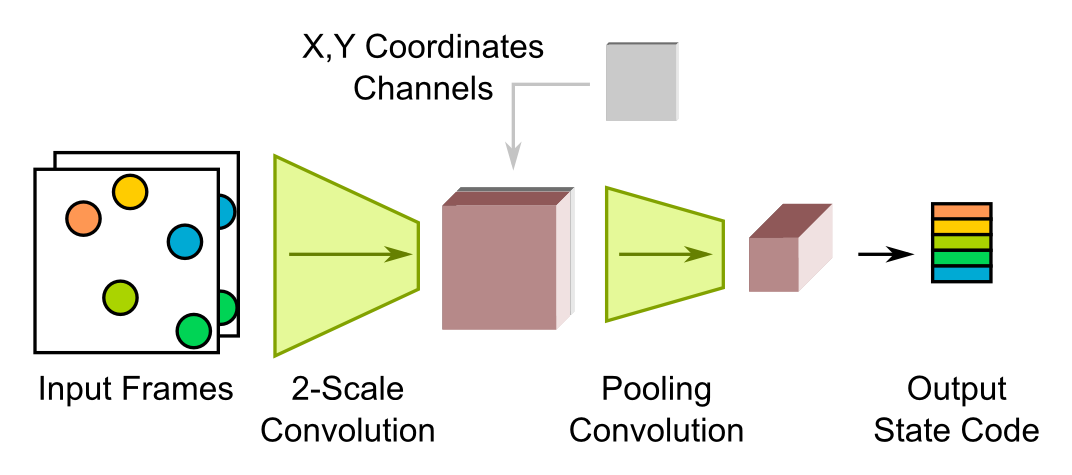}
  \end{subfigure}
  \hspace{0.2in}
  \begin{subfigure}{0.47\textwidth}
  \caption{Interaction Net}
  \label{fig:interaction_net}
  \includegraphics[width=1.0\linewidth]{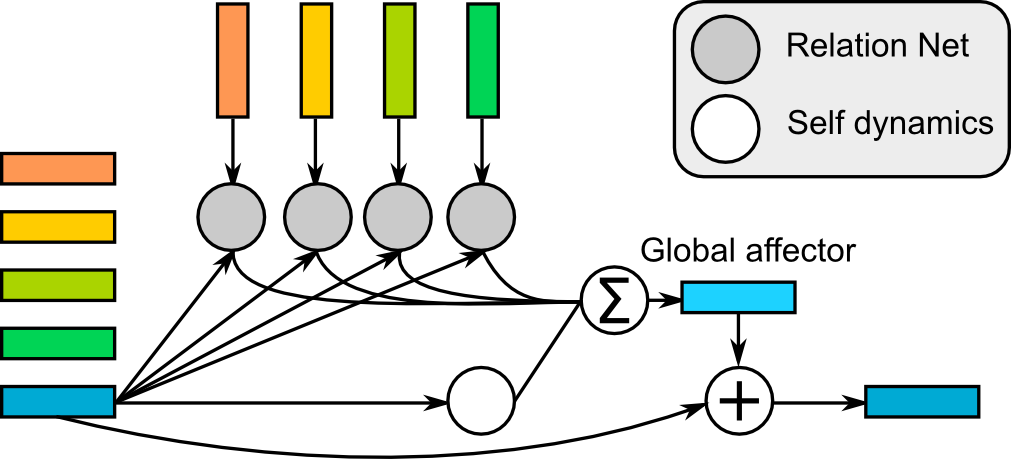}
  \end{subfigure}
  \caption{\textbf{Frame Pair Encoder and Interaction Net.} \textbf{(a)} The frame pair encoder is a CNN which transforms two consecutive input frame into a state code. Important features are the concatenation of coordinate channels before pooling to unit width and height.  The pooled output is reshaped into a state code. \textbf{(b)} An Interaction Net (IN) is used for each temporal offset by the dynamics predictor. For each slot, a relation net is applied to the slot's concatenation with each other slot.  A self-dynamics net is applied to the slot itself. Both of these results are summed and post-processed by the affector to produce the predicted slot.}
  \label{fig:encoder_predictor_cores}
  \vspace{-0.5em}
\end{figure}

The dynamics predictor is a variant of an Interaction Network (IN) \cite{battaglia2016interaction}, a state-to-state physical predictor model summarized in Figure \ref{fig:interaction_net}. The main difference between our predictor and a vanilla IN is aggregation over multiple temporal offsets. Our predictor has a set of temporal offsets (in practice we use $\{1, 2, 4\}$), with one IN core for each. Given an input state code sequence, for each offset $t$ a separate IN core computes a candidate predicted state code from the input state code at index $t$. A slot-wise MLP aggregator transforms the list of candidate state codes into a predicted state code. See the Supplementary Material for further details of the dynamics predictor model.

As with the visual encoder, we explored many dynamics predictor architectures (some of which we compare as baselines below). The temporal offset aggregation of this architecture enhances its power by allowing it to accommodate both fast and slow movements by different objects within a sequence of frames. The factorized representation of INs, which allows efficient learning of interactions even in scenes with many objects, is an important contributor to our predictor architecture's performance.

\subsection{State Decoder}\label{S:decoder}
The state decoder is simply a linear layer with input size $L_{\text{code}}$ and output size 4 (for a position/velocity vector).  This linear layer is applied independently to each slot of the state code. We explored more complicated architectures, but this yielded the best performance. The state decoder is applied to both encoded state codes (for auxiliary encoder loss) and predicted state codes (for prediction loss).

\section{Experiments}\label{S:experiments}
\subsection{Physical Systems Simulations}
We focused on five types of physical systems with high dynamic complexity but low visual complexity, namely 2-dimensional simulations of colored objects on natural-image backgrounds interacting with a variety of forces (see the Supplementary Material for details). In each system the force law is applied pair-wise to all objects and all objects have the same mass and density unless otherwise stated.

\begin{itemize}
    \item \textbf{Spring} Each pair of objects has an invisible spring connection with non-zero equilibrium.  All springs share the same equilibrium and Hooke's constant.
    
    \item \textbf{Gravity} Objects are massive and obey Newton's Law of gravity.
    
    \item \textbf{Billiards} No long-distance forces are present, but the billiards bounce off each other and off the boundaries of the field of vision.
    
    \item \textbf{Magnetic Billiards} All billiards are positively charged, so instead of bouncing, they repel each other according to Coulomb's Law. They still bounce off the boundaries.
    
    \item \textbf{Drift} No forces or interactions of any kind are present. Objects drift forever with their initial velocity.
\end{itemize}

These systems include previously studied gravitational and billiards systems \cite{battaglia2013simulation, agrawal2016learning} with the added challenge of natural image backgrounds.  For example videos of these systems, see the Supplementary Material or visit (\href{https://goo.gl/FD1XX5}{https://goo.gl/FD1XX5})

One limitation of the above systems is that the positions, masses, and radii of all objects are either visually observable in every frame or global constants.  Furthermore, while partial occlusion is allowed, the objects have the same radius so total occlusion never occurs.  In contrast, systems with hidden quantities that influence dynamics abound in the real world.  To mimic this, we explored a few challenging additional systems:
\begin{itemize}
    \item \textbf{Springs with Invisibility}. In each simulation a random object is masked. In this way a model must infer the location of the invisible object from its effects on the other objects.
    \item \textbf{Springs} and \textbf{Billiards with Variable Mass}. In each simulation, each object's radius is randomly generated. This not only causes total occlusion (in the Spring system), but density is held constant, so s model must determine each object's mass from its radius.
\end{itemize}

To simulate each system, we initialized the position and velocity of each ball randomly and uniformly and used a physics engine to simulate the resulting dynamics.  See the Supplementary Material for more details. To generate video data, we rendered the system state on top of a CIFAR-10 natural image background. The background was randomized between simulations and static within each simulation.  Importantly, we rendered the objects with 15-fold anti-aliasing so the visual encoder could learn to distinguish object positions much more finely than pixel resolution, as evident by the visual encoder accuracy described in Section \ref{S:inverse_normalized_loss}.

For each system we generated a dataset with 3 objects and a dataset with 6 objects. Each dataset had a training set of $2.5 \cdot 10^5$ simulations and a test set of $2.5 \cdot 10^4$ simulations, with each simulation 64 frames long. Since we trained on sequences of 14 frames, this ensures we had more than $1 \cdot 10^7$ training samples with distinct dynamics. We added natural image backgrounds online from separate training and testing CIFAR-10 sets, which effectively increased our number of distinct training examples by another factor of 50,000.
\vspace{-0.5em}
\subsection{Baseline Models}
We compared the VIN to a suite of baseline and competitor models, including ablation experiments.  For each model, we performed hyperparameter sweeps across all datasets and choose the hyperparameter set with the lowest average test loss.

The \textbf{Visual RNN} has the same visual encoder as the VIN, but its dynamics predictor is an MLP.  Each state code is flattened before being passed to the dynamics predictor. The dynamics predictor is still treated as a recurrent network with temporal offset aggregation, but the dynamics predictor no longer supports the factorized representation of the IN core.  Without the weight-sharing of the IN, this model is forced to learn the same force law for each pair of objects, which is not scalable as the object number increases.

The \textbf{Visual LSTM} has the same visual encoder as the VIN, but its dynamics predictor is an LSTM \cite{hochreiter1997lstm}. Specifically, the LSTM layer has MLP pre- and post-processors. We also removed the temporal offset aggregation, since the LSTM implicitly integrates temporal information through state updates.  During rollouts (in both training and testing), the output state code from the post-processor MLP is fed into the pre-processor MLP.

The \textbf{VIN Without Relations} is an ablation modification of the VIN.  The only difference between this and the VIN is an omitted relation network in the dynamics predictor cores. Note that there is still ample opportunity to compute relations between objects (both in the visual encoder and the dynamics predictor's temporal offset aggregator), just not specifically through the relation network.

The \textbf{Vision With Ground-Truth Dynamics} model uses a visual encoder and a miniature version of the dynamics predictor to predict not the next-step state but the current-step state (i.e. the state corresponding to the last observed frame). Since this predicts static dynamics, we did not train it on roll-outs. However, when testing, we fed the static state estimation into a ground-truth physics engine to generate rollouts. This model is not a fair comparison to the other models because it does not learn dynamics, only the visual component.  It serves instead as a performance bound dictated by the visual encoder. We normalized our results by the performance of this model, as described in Section \ref{S:results}.

All models described above learn state from \textit{pixels}.  However, we also trained two baselines with privileged information: \textbf{IN from State} and \textbf{LSTM from State} models, which have the IN and LSTM dynamics predictors, but make their predictions directly from state to state. Hence, they do not have a visual encoder but instead have access to the ground truth states for observed frames. These, in combination with the Vision with Ground Truth Dynamics, allowed us to comprehensively test our model in part and in full.

\subsection{Training procedure}\label{S:training}
Our goal was for the models to accurately predict physical dynamics into the future. As shown in Figure \ref{fig:model_framework}, the VIN lends itself well to long-term predictions because the dynamics predictor can be treated as a recurrent net and rolled out on state codes. We trained the model to predict a sequence of 8 unseen future states. Our prediction loss was a normalized weighted sum of the corresponding 8 error terms. The sum was weighted by a discount factor that started at $0.0$ and approached $1.0$ throughout training, so at the start of training the model must only predict the first unseen state and at the end it must predict an average of all 8 future states. Our training loss was the sum of this prediction loss and an auxiliary encoder loss, as indicated in Figure \ref{fig:model_framework}. See the Supplementary Material for full training parameters.

\section{Results}\label{S:results}
Our results show that the VIN predicts dynamics accurately, outperforming baselines on all datasets (see Figures \ref{fig:inv_norm_loss} and \ref{fig:euclidean}). It is scalable, can accommodate forces with a variety of strengths and distance ranges, and can infer visually unobservable quantities (invisible object location) from dynamics. Our model also generates long rollout sequences that are both visually plausible and similar to ground-truth physics, even outperforming state-of-the-art state-to-state models on this measure.

\subsection{Inverse Normalized Loss}\label{S:inverse_normalized_loss}

We evaluated the performance of each model with the Inverse Normalized Loss, defined as $L_{bound} / L_{model}$.  Here $L_{bound}$ is the test loss of the Vision with Ground Truth Dynamics and $L_{model}$ is the test loss of the model in question. We used this metric because the loss itself is not easily interpretable. The Vision with Ground Truth Dynamics produces the best possible predictions given the visual encoder's error, so the Inverse Normalized Loss always lies in $[0, 1]$, where a value closer to $1.0$ indicates better performance.  The visual encoder learned position predictions accurate to within 0.15\% of the framewidth (0.048 times the pixels width), so we have no concerns about the accuracy of the Vision with Ground Truth Dynamics.

\begin{figure}[t]
  \centering
  
  \includegraphics[width=\linewidth]{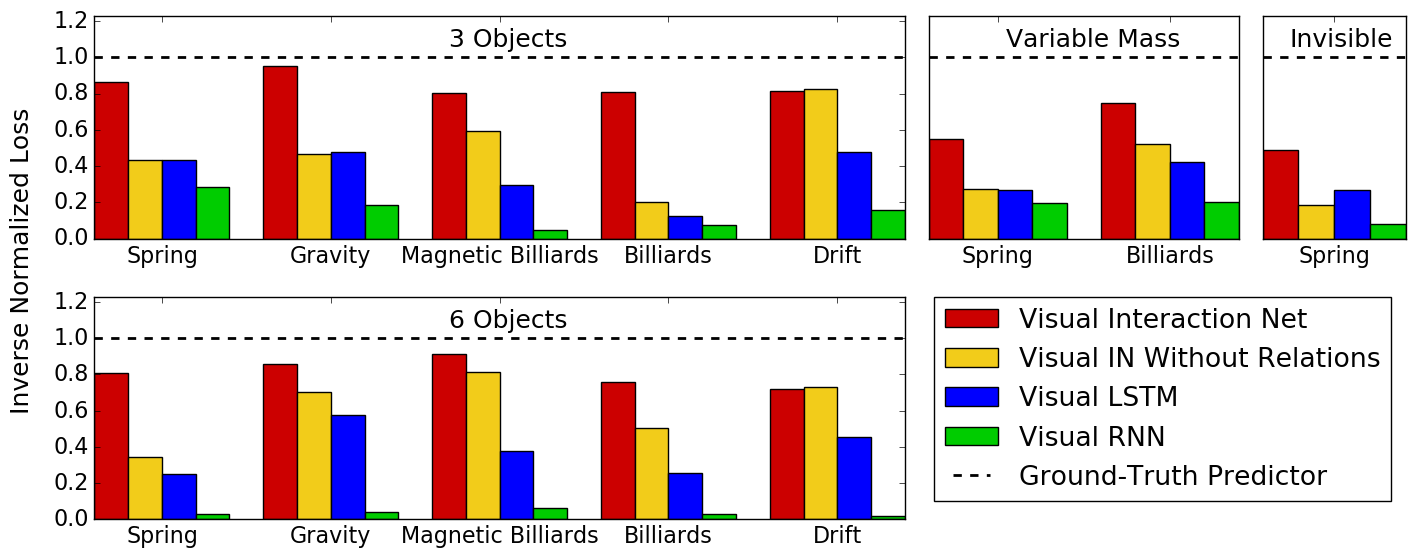}
  
  \caption{\textbf{Performance.} We compare our model's Inverse Normalized Loss to that of the baselines on all datasets. 3-object dataset are on the upper row, and 6-object datasets are on the lower row.  By definition of the Inverse Normalized Loss, all values are in $[0, 1]$ with $1.0$ being the performance of a ground-truth simulator given the visual encoder.  The VIN (red) outperforms every baseline on every dataset (except the VIN Without Relations on Drift, the system with no object interactions).}
  \label{fig:inv_norm_loss}
\end{figure}

Figure \ref{fig:inv_norm_loss} shows the Inverse Normalized Loss on all datasets. The VIN out-performs nearly all baselines on nearly all datasets.  The only baseline with comparable performance is the VIN Without Relations on Drift, which almost perfectly matches the VIN's performance.  This makes sense, because the objects do not interact in the Drift system, so the relation net should be unnecessary.

Of particular note is the performance of the VIN on the invisible dataset (spring system with random invisible object), where its performance is comparable to the fully visible 3-object Spring system. It can locate the invisible object's position to within 4\% of the frame width (1.3 times the pixel width) for the first 8 rollout steps.

\subsection{Euclidean Prediction Error of Rollout Positions}\label{S:rollout_euclidean_deviation}
One important desirable feature of a physical predictor is the ability to extrapolate from a short input video.  We addressed this by comparing performance of all models on long rollout sequences and observing the Euclidean Prediction Error.  To compute the Euclidean Prediction Error from a predicted state and ground-truth state, we calculated the mean over objects of the Euclidean norm between the predicted and true position vectors.

\begin{figure}[t]
  \centering
  
  \begin{subfigure}{0.48\textwidth}
  \caption{Distance Comparison}
  \includegraphics[width=1.0\linewidth]{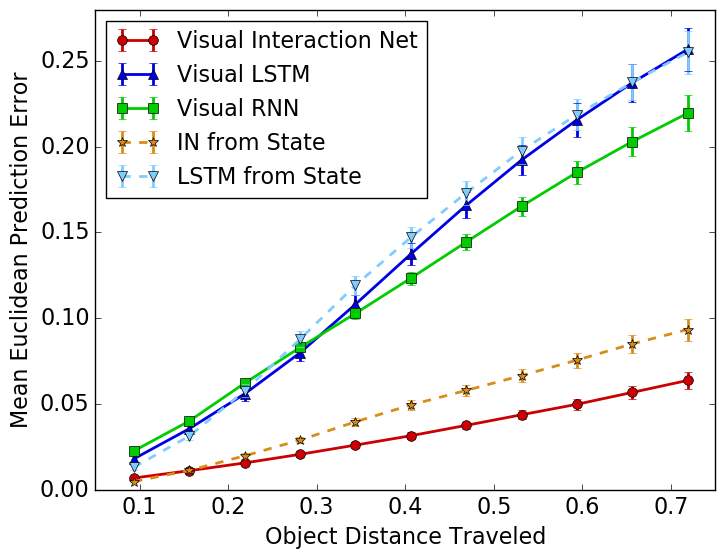}
  \end{subfigure}
  \begin{subfigure}{0.48\textwidth}
  \caption{Time Comparison}
  \includegraphics[width=1.0\linewidth]{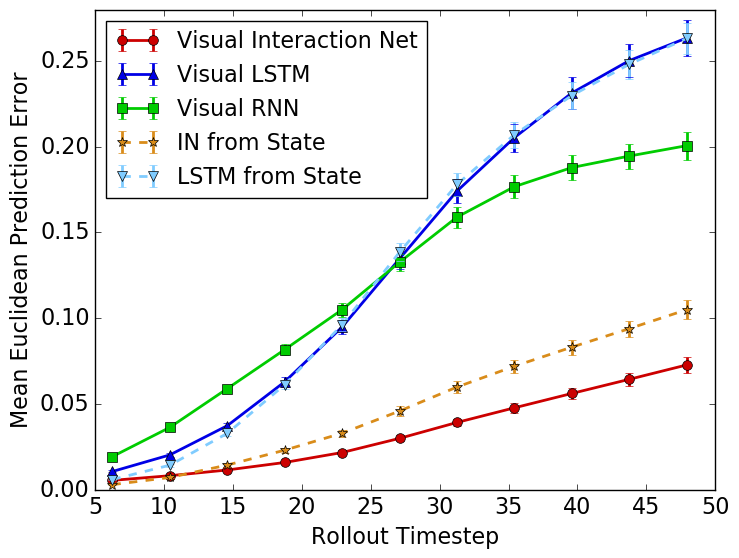}
  \end{subfigure}
  
  \caption{\textbf{Euclidean Prediction Error on 3-object datasets.} We compute the mean over all datasets of the Euclidean Prediction Error from the position predictions of 50-timestep rollouts.  The VIN outperforms all other pixel-to-state models (solid lines) and state-to-state models (dashed lines). Errorbars show 95\% confidence intervals. \textbf{(a)} Mean Euclidean Prediction Error with respect to object distance traveled (measured as a fraction of the frame-width). The VIN is accurate to within 6\% after objects have traversed nearly one framewidth.  \textbf{(b)} Mean Euclidean Prediction Error with respect to timestep. The VIN is accurate to within 7.5\% after 50 timesteps.}
  \label{fig:euclidean}
  \vspace{-0.5em}
\end{figure}

We computed the Euclidean Prediction Error at each step over a 50-timestep rollout sequence. Figure \ref{fig:euclidean} shows the average of this quantity over all 3-object datasets with respect to both timestep and object distance traveled. The VIN out-performs all other models, including the IN from state and LSTM from state even though they have access to privileged information.  This demonstrates the remarkable robustness and generalization power of the VIN.  We believe it outperforms state-to-state models in part because its dynamics predictor must tolerate visual encoder noise during training.  This noise-robustness translates to rollouts, where the dynamics predictor can still produce accurate state-to-state as its predictions deviate from true physical dynamics. The state-to-state methods are not trained on noisy state inputs, so they exhibit poorer generalization. See the Supplementary Material for a dataset-specific quantification of these results.

\subsection{Visualized Rollouts}\label{S:visualized_rollouts}
To qualitatively evaluate the plausibility of the VIN's rollout predictions, we generated videos by rendering the rollout predictions. These are best seen in video format, though we show them in trajectory-trail images here as well. The CIFAR backgrounds made trajectory-trails difficult to see, so we masked the background (just for rendering purposes). Trajectory trails are shown for rollouts between 40 and 60 time steps, depending on the dataset.

We encourage the reader to view the videos at (\href{https://goo.gl/4SSGP0}{https://goo.gl/4SSGP0}). Those include the CIFAR backgrounds and show very long rollouts of up to 200 timesteps, which demonstrate the VIN's extremely realistic rollouts.  We find no reason to doubt that the predictions would continue to be visually realistic (if not exactly tracking the ground-truth simulator) ad infinitum.

\begin{center}
\begin{table}[h!]
\renewcommand{\arraystretch}{0.2}
\begin{tabular}{c c c c c c}
& Spring & Gravity & Magnetic Billiards & Billiards & Drift
\\
\begin{turn}{90} \hspace{0.1in} Example Frame \end{turn}
& \includegraphics[width=0.18\linewidth]{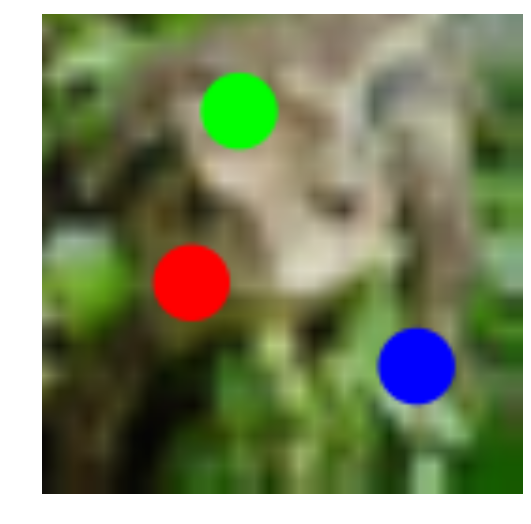}
& \includegraphics[width=0.18\linewidth]{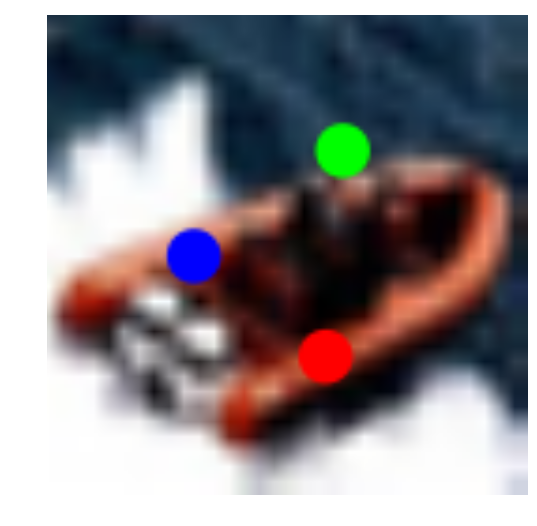}
& \includegraphics[width=0.18\linewidth]{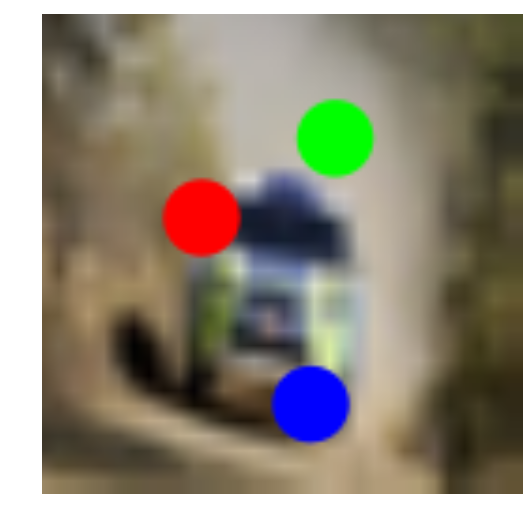}
& \includegraphics[width=0.18\linewidth]{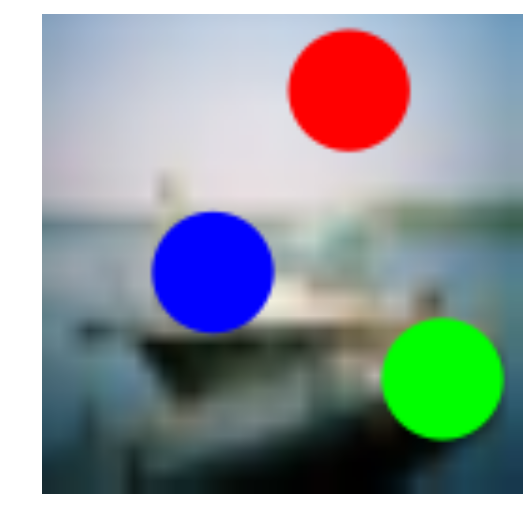}
& \includegraphics[width=0.18\linewidth]{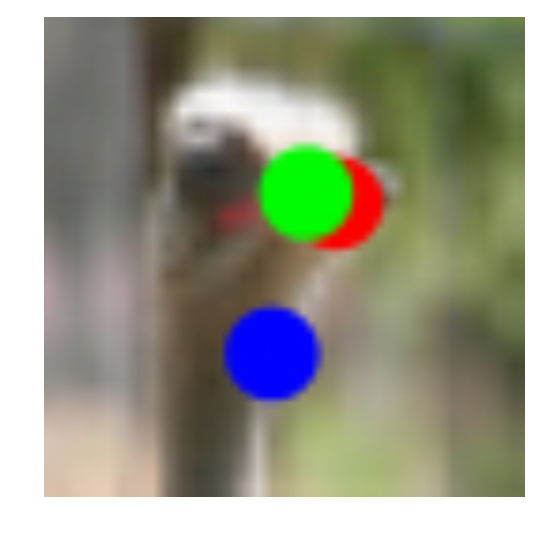}
\\
\begin{turn}{90} \hspace{0.35in} Truth\end{turn}
& \includegraphics[width=0.18\linewidth]{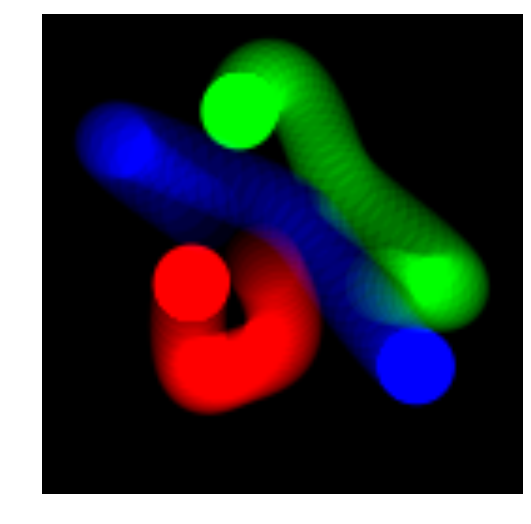}
& \includegraphics[width=0.18\linewidth]{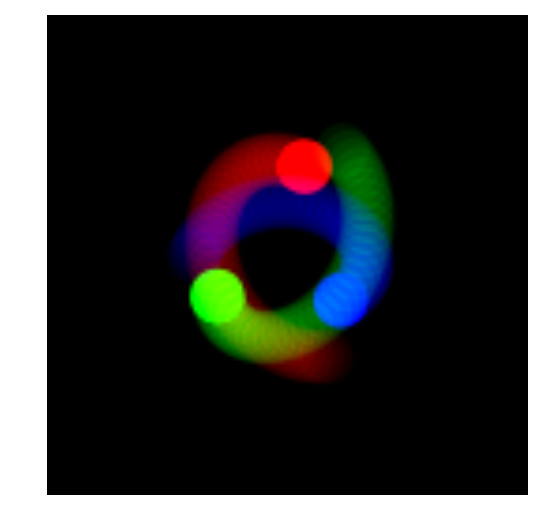}
& \includegraphics[width=0.18\linewidth]{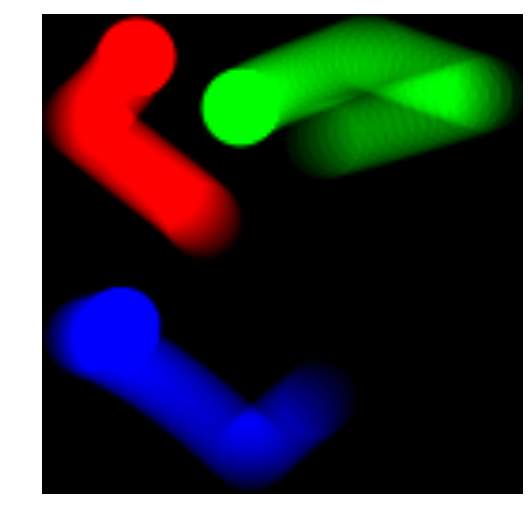}
& \includegraphics[width=0.18\linewidth]{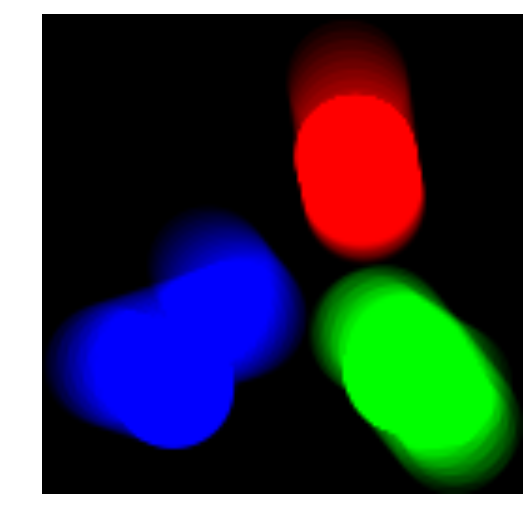}
& \includegraphics[width=0.18\linewidth]{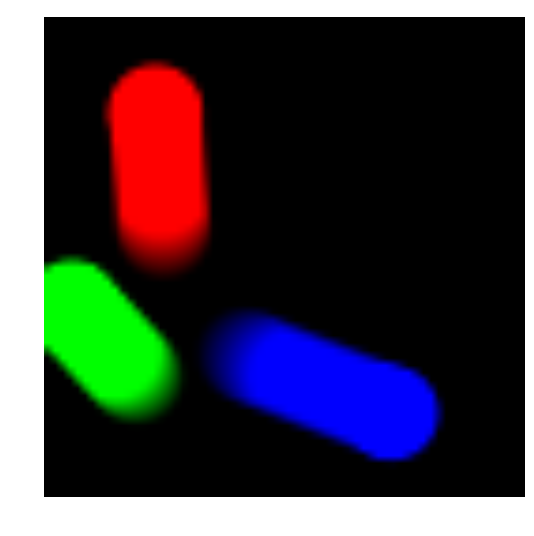}
\\
\begin{turn}{90} \hspace{0.25in} Prediction\end{turn}
& \includegraphics[width=0.18\linewidth]{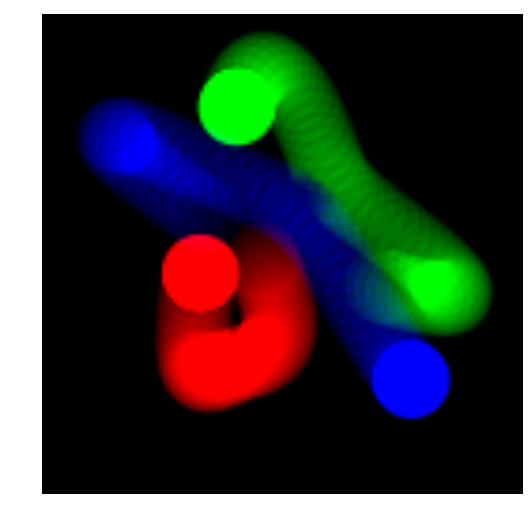}
& \includegraphics[width=0.18\linewidth]{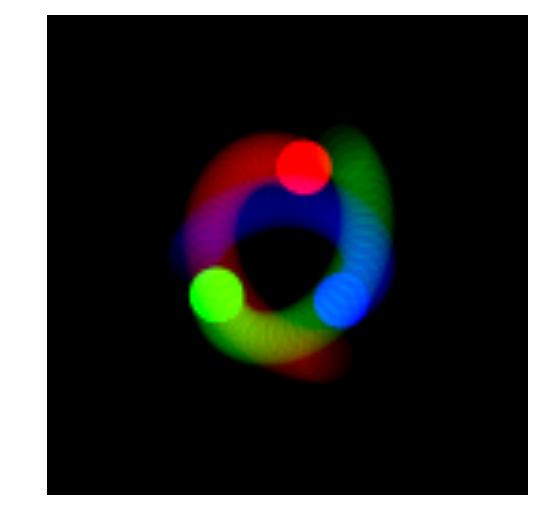}
& \includegraphics[width=0.18\linewidth]{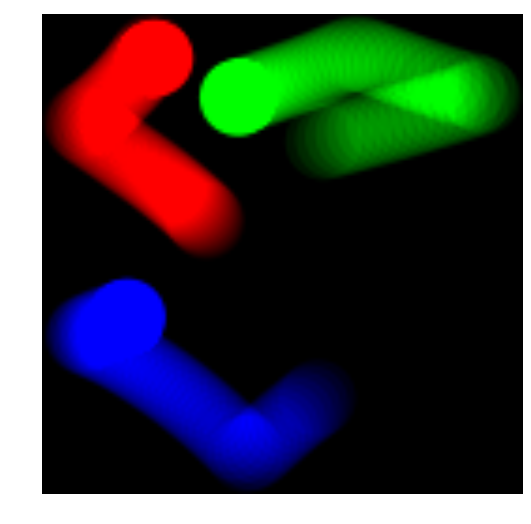}
& \includegraphics[width=0.18\linewidth]{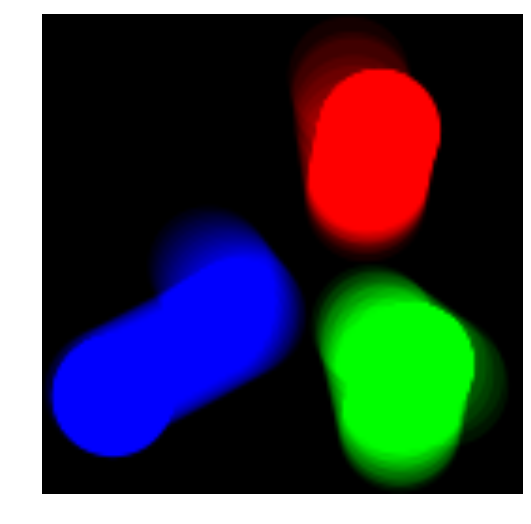}
& \includegraphics[width=0.18\linewidth]{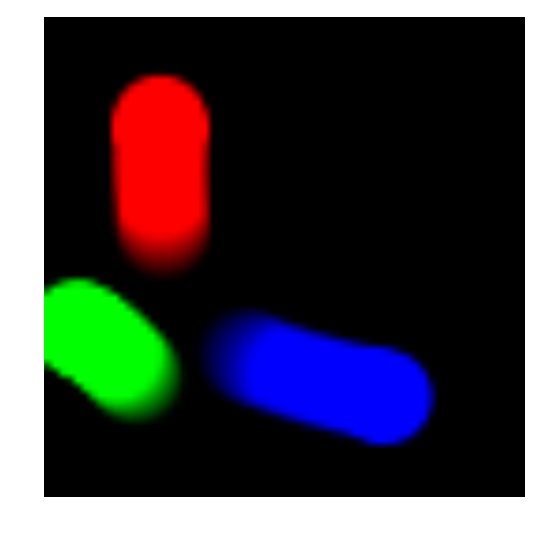}
\\
\vspace{.1in}
\\
\begin{turn}{90} \hspace{0.1in} Example Frame \end{turn}
& \includegraphics[width=0.18\linewidth]{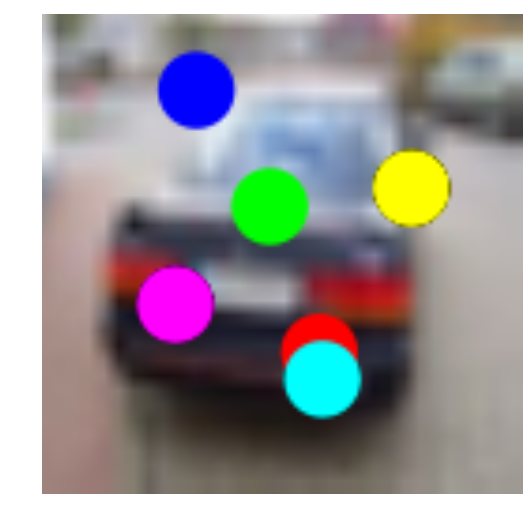}
& \includegraphics[width=0.18\linewidth]{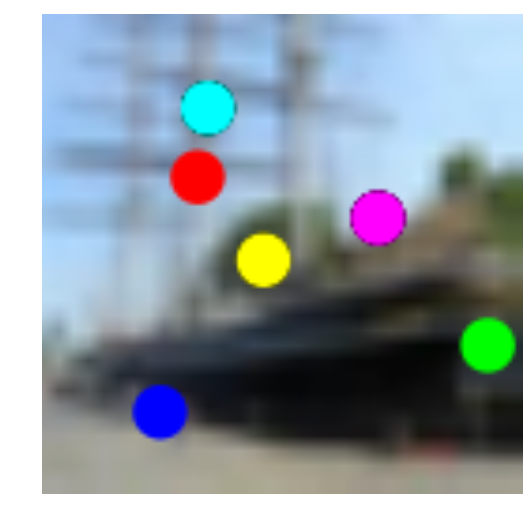}
& \includegraphics[width=0.18\linewidth]{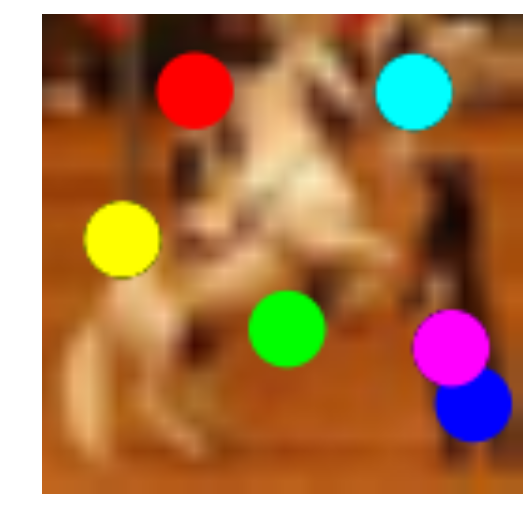}
& \includegraphics[width=0.18\linewidth]{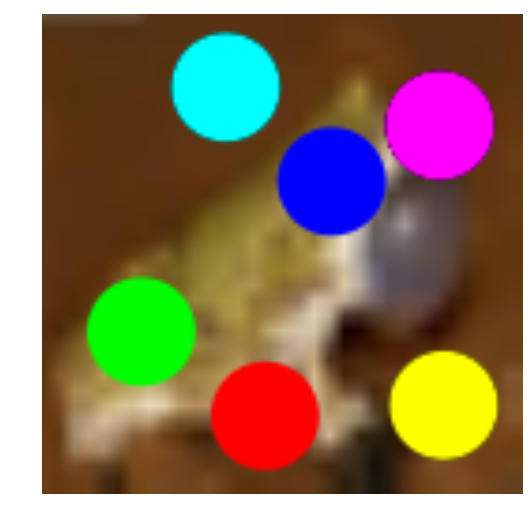}
& \includegraphics[width=0.18\linewidth]{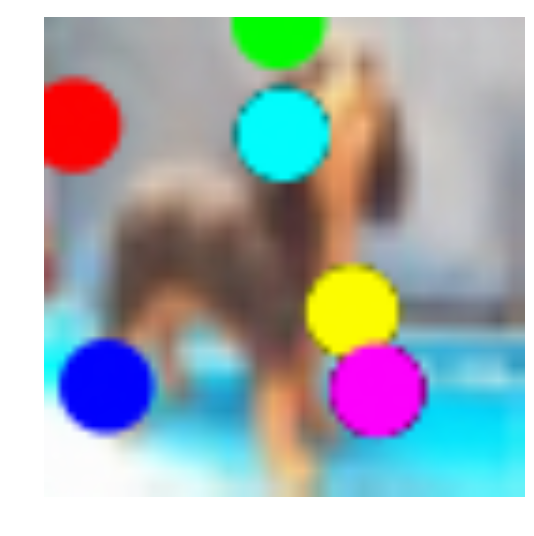}
\\
\begin{turn}{90} \hspace{0.35in} Truth\end{turn}
& \includegraphics[width=0.18\linewidth]{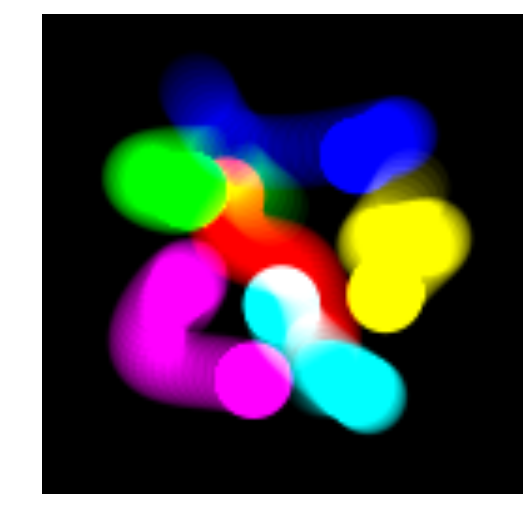}
& \includegraphics[width=0.18\linewidth]{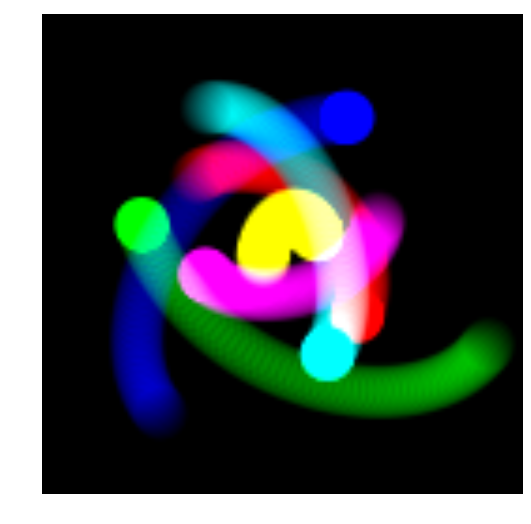}
& \includegraphics[width=0.18\linewidth]{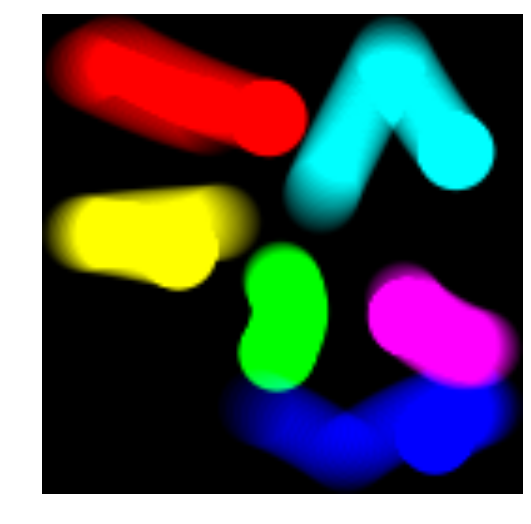}
& \includegraphics[width=0.18\linewidth]{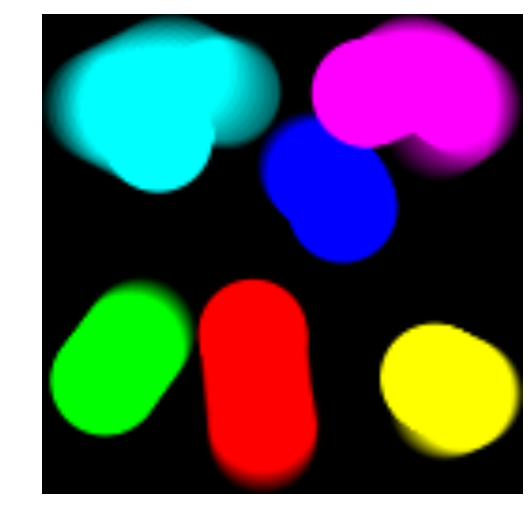}
& \includegraphics[width=0.18\linewidth]{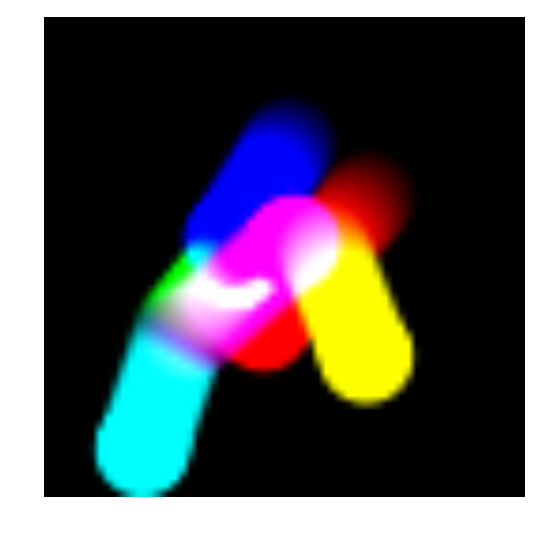}
\\
\begin{turn}{90} \hspace{0.25in} Prediction\end{turn}
& \includegraphics[width=0.18\linewidth]{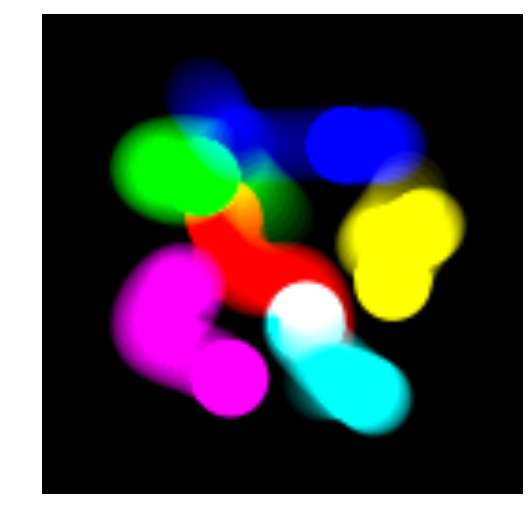}
& \includegraphics[width=0.18\linewidth]{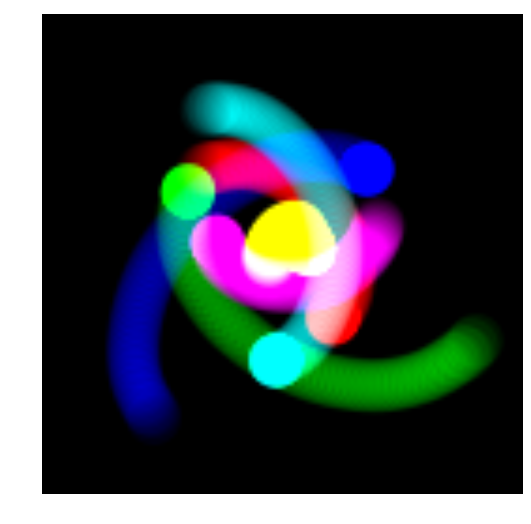}
& \includegraphics[width=0.18\linewidth]{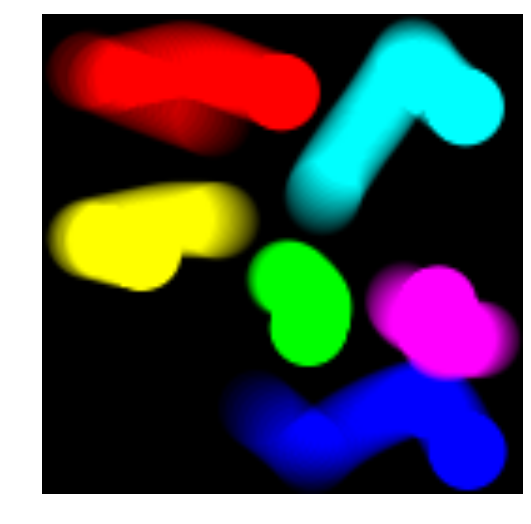}
& \includegraphics[width=0.18\linewidth]{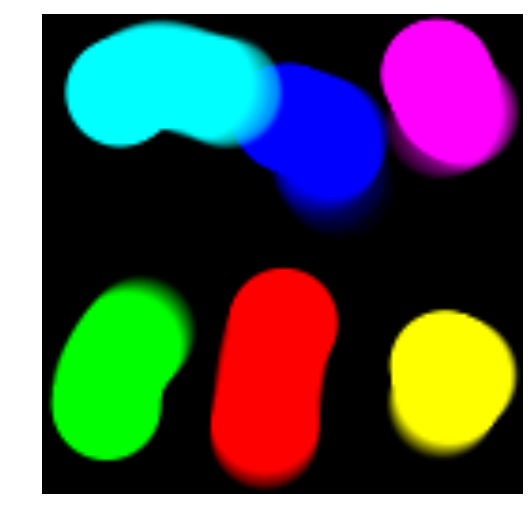}
& \includegraphics[width=0.18\linewidth]{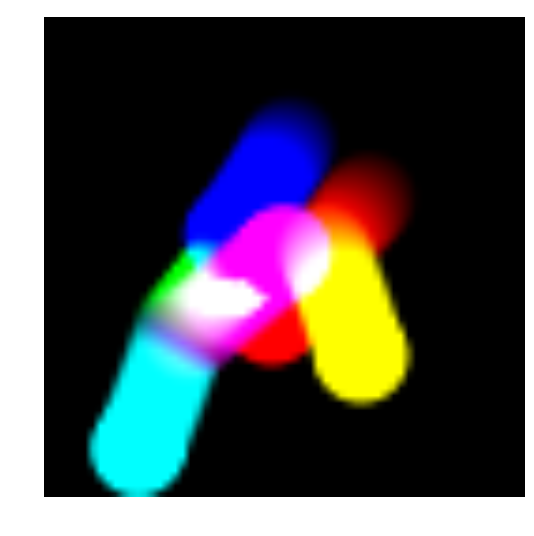}

\end{tabular}\caption{\textbf{Rollout Trajectories.}  For each of our datasets, we show an example frame, an example true future trajectory, and a corresponding predicted rollout trajectory (for 40-60 timesteps, depending on the dataset). This top half shows the 3-object regime and the bottom half shows the 6-object regime. For visual clarity, all objects are rendered at a higher resolution here than in the training input.}
\label{table:rollouts}
\vspace{-2.5em}
\end{table}
\end{center}

\vspace*{1in}

\section{Conclusion}
Here we introduced the Visual Interaction Network and showed how it can infer the physical states of multiple objects from frame sequences and make accurate predictions about their future trajectories. The model uses a CNN-based visual encoder to obtain accurate measurements of object states in the scene. The model also harnesses the prediction abilities and relational computation of Interaction Networks, providing accurate predictions far into the future. We have demonstrated that our model performs well on a variety of physical systems and is robust to visual complexity and partially observable data.

One property of our model is the inherent presence of noise from the visual encoder. In contrast to state-to-state models such as the Interaction Net which operate purely at noiseless state-space, here the dynamic predictor's input is inherently noisy. Even with the relatively clean stimuli we use here (generated synthetically) there is an upper bound to the accuracy achievable with the perceptual front-end. Surprisingly, this noise seemed to confer an advantage because it helped the model learn to overcome temporally compounding errors generated by inaccurate predictions. This is especially notable when doing long term roll outs where we achieve performance which surpasses even a pure state-to-state Interaction Net. Since this dependence on noise would be inherent in any model operating on visual input, we postulate that this is an important feature of \textit{any} prediction model and warrants further research.

Our Visual Interaction Network provides a step toward understanding how representations of objects, relations, and physics can be learned from raw data. This is part of a broader effort toward understanding how perceptual models support physical predictions and how the structure of the physical world influences our representations of sensory input, which will help AI research better capture the powerful object- and relation-based system of reasoning that supports humans' powerful and flexible general intelligence.

\subsubsection*{Acknowledgments}

We would like to thank Max Jaderberg for giving thoughtful feedback on early drafts of this manuscript.

\bibliographystyle{plain}
\small

\bibliography{bibliography}

\begin{thebibliography}{10}

\bibitem{agrawal2016learning}
Pulkit Agrawal, Ashvin Nair, Pieter Abbeel, Jitendra Malik, and Sergey Levine.
\newblock Learning to poke by poking: Experiential learning of intuitive
  physics.
\newblock {\em arXiv preprint arXiv:1606.07419}, 2016.

\bibitem{battaglia2016interaction}
Peter Battaglia, Razvan Pascanu, Matthew Lai, Danilo~Jimenez Rezende, et~al.
\newblock Interaction networks for learning about objects, relations and
  physics.
\newblock In {\em Advances in Neural Information Processing Systems}, pages
  4502--4510, 2016.

\bibitem{battaglia2013simulation}
Peter~W Battaglia, Jessica~B Hamrick, and Joshua~B Tenenbaum.
\newblock Simulation as an engine of physical scene understanding.
\newblock {\em Proceedings of the National Academy of Sciences},
  110(45):18327--18332, 2013.

\bibitem{bhat2002computing}
Kiran Bhat, Steven Seitz, Jovan Popovi{\'c}, and Pradeep Khosla.
\newblock Computing the physical parameters of rigid-body motion from video.
\newblock {\em Computer Vision—ECCV 2002}, pages 551--565, 2002.

\bibitem{bhattacharyya2016long}
Apratim Bhattacharyya, Mateusz Malinowski, Bernt Schiele, and Mario Fritz.
\newblock Long-term image boundary extrapolation.
\newblock {\em arXiv preprint arXiv:1611.08841}, 2016.

\bibitem{brubaker2009estimating}
Marcus~A Brubaker, Leonid Sigal, and David~J Fleet.
\newblock Estimating contact dynamics.
\newblock In {\em Computer Vision, 2009 IEEE 12th International Conference on},
  pages 2389--2396. IEEE, 2009.

\bibitem{chang2016compositional}
Michael~B Chang, Tomer Ullman, Antonio Torralba, and Joshua~B Tenenbaum.
\newblock A compositional object-based approach to learning physical dynamics.
\newblock {\em arXiv preprint arXiv:1612.00341}, 2016.

\bibitem{ehrhardt2017learning}
Sebastien Ehrhardt, Aron Monszpart, Niloy~J Mitra, and Andrea Vedaldi.
\newblock Learning a physical long-term predictor.
\newblock {\em arXiv preprint arXiv:1703.00247}, 2017.

\bibitem{fragkiadaki2015learning}
Katerina Fragkiadaki, Pulkit Agrawal, Sergey Levine, and Jitendra Malik.
\newblock Learning visual predictive models of physics for playing billiards.
\newblock {\em arXiv preprint arXiv:1511.07404}, 2015.

\bibitem{gerstenberg2012noisy}
Tobias Gerstenberg, Noah Goodman, David~A Lagnado, and Joshua~B Tenenbaum.
\newblock Noisy newtons: Unifying process and dependency accounts of causal
  attribution.
\newblock In {\em In proceedings of the 34th}. Citeseer, 2012.

\bibitem{gerstenberg2014counterfactual}
Tobias Gerstenberg, Noah Goodman, David~A Lagnado, and Joshua~B Tenenbaum.
\newblock From counterfactual simulation to causal judgment.
\newblock In {\em CogSci}, 2014.

\bibitem{grzeszczuk1998neuroanimator}
Radek Grzeszczuk, Demetri Terzopoulos, and Geoffrey Hinton.
\newblock Neuroanimator: Fast neural network emulation and control of
  physics-based models.
\newblock In {\em Proceedings of the 25th annual conference on Computer
  graphics and interactive techniques}, pages 9--20. ACM, 1998.

\bibitem{hamrick2016inferring}
Jessica~B Hamrick, Peter~W Battaglia, Thomas~L Griffiths, and Joshua~B
  Tenenbaum.
\newblock Inferring mass in complex scenes by mental simulation.
\newblock {\em Cognition}, 157:61--76, 2016.

\bibitem{hochreiter1997lstm}
Sepp Hochreiter and Jurgen Schmidhuber.
\newblock Long short-term memory.
\newblock {\em Neural Computation}, 9(8):1735--1780, 1997.

\bibitem{ladicky2015data}
Lubor Ladicky, SoHyeon Jeong, Barbara Solenthaler, Marc Pollefeys, Markus
  Gross, et~al.
\newblock Data-driven fluid simulations using regression forests.
\newblock {\em ACM Transactions on Graphics (TOG)}, 34(6):199, 2015.

\bibitem{lecun2015deep}
Yann LeCun, Yoshua Bengio, and Geoffrey Hinton.
\newblock Deep learning.
\newblock {\em Nature}, 521(7553):436--444, 2015.

\bibitem{lerer2016learning}
Adam Lerer, Sam Gross, and Rob Fergus.
\newblock Learning physical intuition of block towers by example.
\newblock {\em arXiv preprint arXiv:1603.01312}, 2016.

\bibitem{li2016fall}
Wenbin Li, Seyedmajid Azimi, Ale{\v{s}} Leonardis, and Mario Fritz.
\newblock To fall or not to fall: A visual approach to physical stability
  prediction.
\newblock {\em arXiv preprint arXiv:1604.00066}, 2016.

\bibitem{mottaghi2016newtonian}
Roozbeh Mottaghi, Hessam Bagherinezhad, Mohammad Rastegari, and Ali Farhadi.
\newblock Newtonian scene understanding: Unfolding the dynamics of objects in
  static images.
\newblock In {\em Proceedings of the IEEE Conference on Computer Vision and
  Pattern Recognition}, pages 3521--3529, 2016.

\bibitem{mottaghi2016happens}
Roozbeh Mottaghi, Mohammad Rastegari, Abhinav Gupta, and Ali Farhadi.
\newblock “what happens if...” learning to predict the effect of forces in
  images.
\newblock In {\em European Conference on Computer Vision}, pages 269--285.
  Springer, 2016.

\bibitem{spelke2007core}
Elizabeth~S Spelke and Katherine~D Kinzler.
\newblock Core knowledge.
\newblock {\em Developmental science}, 10(1):89--96, 2007.

\bibitem{stewart2016label}
Russell Stewart and Stefano Ermon.
\newblock Label-free supervision of neural networks with physics and domain
  knowledge.
\newblock {\em arXiv preprint arXiv:1609.05566}, 2016.

\bibitem{winograd1971procedures}
Terry Winograd.
\newblock Procedures as a representation for data in a computer program for
  understanding natural language.
\newblock Technical report, DTIC Document, 1971.

\bibitem{winston1970learning}
Patrick~H Winston.
\newblock Learning structural descriptions from examples.
\newblock 1970.

\bibitem{wu2016physics}
Jiajun Wu, Joseph~J Lim, Hongyi Zhang, Joshua~B Tenenbaum, and William~T
  Freeman13.
\newblock Physics 101: Learning physical object properties from unlabeled
  videos.
\newblock {\em psychological science}, 13(3):89--94, 2016.

\bibitem{wu2015galileo}
Jiajun Wu, Ilker Yildirim, Joseph~J Lim, Bill Freeman, and Josh Tenenbaum.
\newblock Galileo: Perceiving physical object properties by integrating a
  physics engine with deep learning.
\newblock In {\em Advances in neural information processing systems}, pages
  127--135, 2015.

\end{thebibliography}

\newpage

\centerline{\LARGE{\textsc{Supplementary Material}}}

\section{Supplementary Videos}\label{S:supp_video}

We provide videos showing sample VIN rollout sequences and dataset examples. In all videos, for visual clarity the objects are rendered at a higher resolution than they are in the input data.

\subsection{Sample VIN Rollout Videos}

See the videos at

\href{https://goo.gl/4SSGP0}{https://goo.gl/4SSGP0}

These show rendered VIN rollout position predictions compared to the ground-truth (unobserved) system simulation for 3-object and 6-object datasets of all force systems.  In each video, the VIN rollout is on the right-hand side and the ground-truth on the left-hand side.  Rollouts are 200 steps long for 3-object systems and 100 steps long for 6-object systems (except for Drift, where we use 35 steps for all rollouts to ensure objects don't drift out of view).  The VIN rollout tracks the ground truth quite closely for most datasets.  Three rollout examples are provided for each dataset.

\subsection{Dataset Examples}

See the videos at

\href{https://goo.gl/FD1XX5}{https://goo.gl/FD1XX5}

These show examples of the 3-object and 6-object datasets of all force systems.  The VIN and all baselines receive 6 frames of these as input during training and rollouts.

\section{Training Parameters}

We use the following training parameters for all models:
\begin{itemize}
    \item Training steps:  $5\cdot 10^5$
    \item Batch Size: 4
    \item Gradient Descent Optimizer:  Adam, learning rate $5\cdot 10^{-4}e^{-t / \alpha}$ where $\alpha = 1.5\cdot 10^5$ and $t$ is the training step.
    \item Rollout frame temporal discount (factor by which future frames are weighted less in the loss): $1 - \gamma$ with $\gamma = e^{-t / \beta}$ where $\beta = 2.5\cdot 10^4$ and $t$ is the training step.
\end{itemize}

\section{VIN Model Details}\label{S:supp_model}

\subsection{Visual Encoder}\label{S:supp_visual_encoder}

See the main text for schematics and high-level summary descriptions of all model components. Here we describe parameters and details.

The \textbf{visual encoder} takes a sequence of three images as input and outputs a state code.  It's sequence of operations on frames $[F_1, F_2, F_3]$ is as follows:
\begin{itemize}
    \item Apply an image pair encoder (described below) to $[F_1, F_2]$ and $[F_2, F_3]$, obtaining $S_1$ and $S_2$.  These are length-32 vectors.
    \item Apply a shared linear layer to convert $S_1$ and $S_2$ to tensors of shape $N_{object}\times 64$.  Here $N_{object}$ is the number of objects in the scene, and 64 is the length of each state code slot.
    \item Concatenate $S_1$ and $S_2$ in a slot-wise manner, obtaining a single tensor $S$ of shape $N_{object}\times 128$.
    \item Apply a shared MLP with one hidden layer of size $64$ and a length-$64$ output layer to each slot of $S$. The result is the encoded state code.
\end{itemize}

The Image Pair Encoder takes two images as input and outputs a candidate state code. It's sequence of operations on frames $[F_1, F_2]$ is as follows:
\begin{itemize}
    \item Stack $F_1$ and $F_2$ along their color-channel dimension.
    \item Independently apply two 2-layer convolutional nets, one with kernel size 10 and 4 channels and the other with kernel size 3 and 16 channels. Both are padded to preserve the input size. Stack the outputs of these convolutions along the channel dimension.
    \item Apply a 2-layer size-preserving convolutional net with 16 channels and kernel-size 3.
    \item Inject two constant coordinate channels, representing the x- and y-coordinates of the feature matrix. These two channels are a meshgrid with min value 0 and max value 1.
    \item Convolve to unit height and width with alternating convolutional and $2 \times 2$ max-pooling layers.  The convolutional layers are size-preserving and have kernel size 3.  In total, there are 5 each of convolutional and max-pooling layers. The first three layers have 16 channels, and the last two have 32 channels. Flatten the result into a 32-length vector. This is the image pair encoder's output.
\end{itemize}

\subsection{Dynamics Predictor}\label{S:supp_predictor}

The \textbf{dynamics predictor} takes a sequence of 4 consecutive state codes $[S_1, ..., S_4]$ and outputs a predicted state code $S^{pred}$, as follows:
\begin{itemize}
    \item Temporal offsets are ${1, 2, 4}$, so we have IN cores $C_1, C_2, C_4$.  Since the temporal offset indexing goes back in time, we apply $C_4$ to $S_1$, $C_2$ to $S_3$, and $C_1$ to $S_4$.  Let $S^{candidate}_1, S^{candidate}_3, S^{candidate}_4$ denote the outputs.
    
    \item Apply a shared slot-wise MLP aggregator with sizes $[32, 64]$ to the concatenation of $S_i^{1, 3, 4}$ for each $i\in \{1, ..., N_{object}\}$. The resulting state code is the dynamics predictor's output.
\end{itemize}

The \textbf{Interaction Net} core takes a state code $[M_i]_{1\leq i\leq N_{object}}$ as input and outputs a candidate state code, as follows:
\begin{itemize}
    \item Apply a Self-Dynamics MLP with sizes $[64, 64]$ to each slot $M_i$. Let $[M^{self}_i]_{1\leq i\leq N_{object}}$ denote these.
    
    \item Apply a Relation MLP with sizes $[64, 64, 64]$ to the concatenation of each pair of distinct slots. Let $[M^{rel}_{ij}]_{1\leq i \neq j\leq N_{object}}$ denote the outputs.
    
    \item Sum for each slot the quantities computed so far, to produce an updated slot.  Specifically, let $M^{update}_i = M^{self}_i + \sum_j M^{rel}_{ij}$.
    
    \item Apply an Affector MLP with sizes $[64, 64, 64]$ to each $M^{update}_i$, yielding $M^{affect}_i$.
    
    \item For each slot, apply a shared MLP with sizes $[32, 64]$ to the concatenation of $M_i$ and $M^{affect}_i$.  The resulting state code is the output.
    
\end{itemize}

The \textbf{IN from State} model uses the same dynamics predictor, but no encoder (it is given the position/velocity vectors for directly).

\subsection{RNN Models}\label{S:supp_rnn}

The \textbf{Visual RNN} model uses the same parameters for the visual encoder as the VIN. The RNN dynamics predictor has a single MLP core with sizes $[64, 64, 64, 64, 64]$ and the same temporal offset of $\{1, 2, 4\}$ and slot-wise MLP aggregator with sizes $[32, 64]$ as the VIN.

\subsection{LSTM}\label{S:supp_lstm}

The \textbf{Visual LSTM} model uses the same parameters for the visual encoder as the VIN. The LSTM dynamics predictor has a single LSTM/MLP core consisting of a pre-processor MLP with sizes $[64, 64]$, an LSTM with $128$ hidden units, and a post-processor MLP with sizes $32, 32$. This is the core of an temporal offset-aggregating MLP with sizes $[32, 64]$ and temporal offsets ${1, 2, 4}$.

The \textbf{LSTM from State} model uses the LSTM dynamics predictor on position/velocity states.

\section{Datasets}

\subsection{Physical Systems}

We simulated each physical system with Newton's Method and internal simulation timestep small enough that there was no visual distinction after 300 frames when using the RK4 method. We use the specific force laws below:

\begin{itemize}
    \item \textbf{Spring} 
    A pair of objects at positions $\vec p_i$ and $\vec p_i$ obey Hooke's law
    \begin{align}
        \vec F_{ij} &= -\kappa \vec d_{ij} - \varepsilon \frac{\vec d_{ij}}{|\vec d_{ij}|}\notag
    \end{align}
    where $F_{ij}$ is the force component on object $j$ from object $i$. Here $\vec d_{ij} = \vec p_i - \vec p_j$ is the displacement between the objects, $\kappa$ is the spring constant, and $\varepsilon$ is the equilibrium. We use $\varepsilon=0.45$
    
    \item \textbf{Gravity}
    A the pair of objects with masses $m_i$, $m_j$ obey Newton's Law
    \begin{align}
        \vec F_{ij} &= -G \frac{m_i m_j \vec d_{ij}}{|\vec d_{ij}|^3}\notag
    \end{align}
    where $G$ is the gravitational constant. In practice, we upper-bounded the gravitational force to avoid instability due to the "slingshot" effect when two objects pass extremely close to each other. To further prevent objects from drifting out of view, we also applied a weak attraction towards the center of the field of view. The system effectually operates within a parabolic bowl.  
    
    \item \textbf{Billiards}
    A pair of balls only interact when they touch, in which case they bounce off of each other instantaneously and with total elasticity. The bounces conserve kinetic energy and total momentum, as if the objects are perfect billiard balls. In addition, the balls bounce off the edges of the field of view.
    
    \item \textbf{Magnetic Billiards}
    A pair of objects with charges $q_i$, $q_j$ obey Coulomb's law
    \begin{align}
        F_{ij} &= k\frac{q_1 q_2 \vec d_{ij}}{|\vec d_{ij}|^3}\notag
    \end{align}
    where $k$ is Coulomb's constant. In addition, the balls bounce off the edges of the field of view as in the Billiards system.
    
    \item \textbf{Drift}
    In this system there are no forces, so the objects simply drift with their initial velocity. We terminate all simulations before the objects completely exit the frame, though bound the initial positions and velocities so that this never occurs before 32 timesteps.
\end{itemize}

Unspecified parameters $\kappa$, $G$, and $k$ were tuned with the render stride for each dataset and object number to make the object velocities look reasonable.

We initialize each object's initial position uniformly within a centered box of width $0.8$ times the framewidth. We initialize each object's velocity uniformly at random, except for Gravity, where we initialize each object's velocity as the counter-clockwise vector tangent with respect to the center of the frame, then add a small random vector (this was necessary to ensure stability of the system).

For the unbounded systems (Gravity, Springs, and Drift), after the velocities are initialized we enforce net zero momentum by subtracting an appropriate vector from each ball's initial velocity. For Gravity and Springs, we also center the objects' positions so that the center of mass lies in the center of the frame. These measures ensures the entire system does not drift out of view.

For all systems except Drift we apply a weak frictional force (linearly proportional to each ball's area), to ensure that any accumulation of numerical inaccuracies does not cause instability in any systems, even after many hundreds of timesteps.

We render each system as a 32 $\times$ 32 RGB video in front of a CIFAR10 natural image background. For systems that allow occlusion (every system except Billiards), we use an foreground/background ordering of the balls by color, and this ordering is fixed for the entire dataset.

\subsection{Numerical Results}\label{S:supp_results_tables}

In Tables \ref{table:euclidean_8} and \ref{table:euclidean_50} we show values of the Mean Euclidean Prediction error on all models and all datasets after 8 and 50 rollout steps, respectively. These values numerically represent time-slices of Figure 4 in the main text, partitioned by dataset.

\begin{center}
\begin{table}[htb]
\begin{tabular}{| c | c c c | c c |}
\multicolumn{1}{c |}{} &
\multicolumn{3}{c |}{pixel-to-state models} &
\multicolumn{2}{c |}{state-to-state models} \\
\hline
3-object datasets & VIN & Visual LSTM & Visual RNN & VIN from State & LSTM from State \\
\hline
Spring & 1.831 & 3.272 & 1.646 & 0.426 & 1.844 \\
\hline
Gravity & 1.288 & 1.572 & 1.194 & 0.146 & 0.191 \\
\hline
Magnetic Billiards & 1.878 & 2.911 & 1.792 & 0.454 & 1.863 \\
\hline
Billiards & 1.600 & 2.752 & 1.391 & 0.942 & 2.507 \\
\hline
Drift & 2.920 & 3.663 & 2.474 & 0.0017 & 0.0052 \\
\hline
& & & & &\\
\hline
6-object datasets & VIN & Visual LSTM & Visual RNN & VIN from State & LSTM from State \\
\hline
Spring & 0.608 & 0.858 & 0.565 & 0.235 & 0.324 \\
\hline
Gravity & 0.422 & 0.597 & 0.416 & 0.092 & 0.157 \\
\hline
Magnetic Billiards & 0.836 & 1.374 & 0.750 & 0.349 & 0.791 \\
\hline
Billiards & 1.022 & 2.582 & 0.918 & 0.817 & 1.919 \\
\hline
Drift & 0.831 & 1.083 & 0.749 & 0.0025 & 0.0069 \\
\hline
\end{tabular}\caption{\textbf{Mean Euclidean Prediction Error for 8-Step Rollouts.} These values show the Mean Euclidean Prediction Error on the length-8 test rollouts.  All values are scaled by 100, so they show a geometric deviation as a percentage of the frame width. Our model outperforms all pixel-to-state baselines on all datsets, and also outperforms the LSTM state-to-state baseline on some datasets. We believe the unexpectedly high values on Drift result from objects nearly drifting out of view before simulations are terminated. We also believe the lower values on 6-object datasets is a function of the slower velocity of those datasets.}
\end{table}\label{table:euclidean_8}
\end{center}

\begin{center}
\begin{table}[htb]
\begin{tabular}{| c | c c c | c c |}
\multicolumn{1}{c |}{} &
\multicolumn{5}{c |}{Euclidean deviation after 50 simulation timesteps} \\
\hline
& Our Model & Visual LSTM & Visual RNN & Our Predictor & LSTM Predictor \\
\hline\hline
Spring & 0.046 & 0.249 & 0.157 & 0.063 & 0.324 \\
\hline
Gravity & 0.008 & 0.048 & 0.043 & 0.013 & 0.081 \\
\hline
Magnetic Billiards & 0.111 & 0.398 & 0.314 & 0.179 & 0.332 \\
\hline
Billiards & 0.151 & 0.391 & 0.308 & 0.199 & 0.348 \\
\hline
\multicolumn{1}{c |}{} &
\multicolumn{5}{c |}{Euclidean deviation per object after full frame width is travelled} \\
\hline
& Our Model & Visual LSTM & Visual RNN & Our Predictor & LSTM Predictor \\
\hline\hline
Spring & 0.069 & 0.304 & 0.213 & 0.091 & 0.360 \\
\hline
Gravity & 0.009 & 0.038 & 0.038 & 0.010 & 0.057 \\
\hline
Magnetic Billiards & 0.118 & 0.417 & 0.455 & 0.165 & 0.354 \\
\hline
Billiards & 0.179 & 0.470 & 0.395 & 0.223 & 0.411 \\
\hline
\end{tabular}\caption{\textbf{Mean Euclidean Prediction Error for 50-Step Rollouts.} These values show the Mean Euclidean Prediction Error on length-50 test rollouts.  All values are scaled by 100, so they show a geometric deviation as a percentage of the frame width. Our model out-performs all other models, including state-to-state models.}
\end{table}\label{table:euclidean_50}
\end{center}

\end{document}